\documentclass[10pt,twocolumn,letterpaper]{article}

\usepackage{cvpr}
\usepackage{times}
\usepackage{epsfig}
\usepackage{graphicx}
\usepackage{amsmath}
\usepackage{amssymb}
\usepackage{multirow}
\usepackage[para]{threeparttable}

\usepackage[breaklinks=true,bookmarks=false]{hyperref}

\cvprfinalcopy 

\def\cvprPaperID{****} 

\begin{document}

\title{Neural Stain Normalization and Unsupervised Classification\\ of Cell Nuclei in Histopathological Breast Cancer Images}

\author{Edwin Yuan\\
Department of Applied Physics\\
Stanford Univeristy, CA, USA\\
{\tt\small edyuan@stanford.edu}
\and
Junkyo Suh\\
Department of Electrical Engineering\\
Stanford Univeristy, CA, USA\\
{\tt\small suhjk@stanford.edu}
}

\maketitle

\begin{abstract}
In this paper, we develop a complete pipeline for  stain normalization, segmentation, and classification of nuclei in hematoxylin and eosin (H\&E) stained breast cancer histopathology images. In the first step, we use a CNN-based stain transfer technique to normalize the staining characteristics of (H\&E) images. We then train a neural network to segment images of nuclei from the H\&E images. Finally, we train an Information Maximizing Generative Adversarial Network (InfoGAN) to learn visual representations of different types of nuclei and classify them in an entirely unsupervised manner. The results show that our proposed CNN stain normalization yields improved visual similarity and cell segmentation performance compared to the conventional SVD-based stain normalization method. In the final step of our pipeline, we demonstrate the ability to perform fully unsupervised clustering of various breast histopathology cell types based on morphological and color attributes. In addition, we quantitatively evaluate our neural network - based techniques against various quantitative metrics to validate the effectiveness of our pipeline.
\end{abstract}

\section{Introduction}

Breast cancer is the second leading cause of cancer death among women, but early diagnosis significantly increases treatment success.  Ductal carcinoma, which 80\% of breast cancers begin as, has survival rates of nearly 100\% if the growing tumor is resected before it has a chance to spread. The foremost technique for diagnosis is based on a specialist's evaluation of hematoxylin and eosin (H\&E) stained breast histopathology images, which are thin 5 micron slices of tissue that have been resected from the patient, and give a 2d representation of the cellular and stromal activity inside the tissue.   However, the sheer amount of H\&E data generated from even a small block of tissue is enormous, and this poses intrinsic challenges to human pathologists.   Firstly, the H\&E sections contain biological information that exists at many scales and across a very large spatial area, from cell-to-cell interaction (microns), to the dynamics of a group of cells (tens of microns), to the organization of the tissue as a whole (millimeters).  Humans are instead naturally suited to rationalizing single events in a cause and effect manner, using concepts derived from an internal mental model.  The typical trained human pathologist resorts to sampling the H\&E data, focusing first on selected spatial locations, and then gathering an impression of the overall state of the tissue, in order to finally use a set of knowledge-based and experience-based heuristics that guide him or her towards a diagnosis.

\begin{figure}[t]
\begin{center}
   \includegraphics[ scale=0.32]{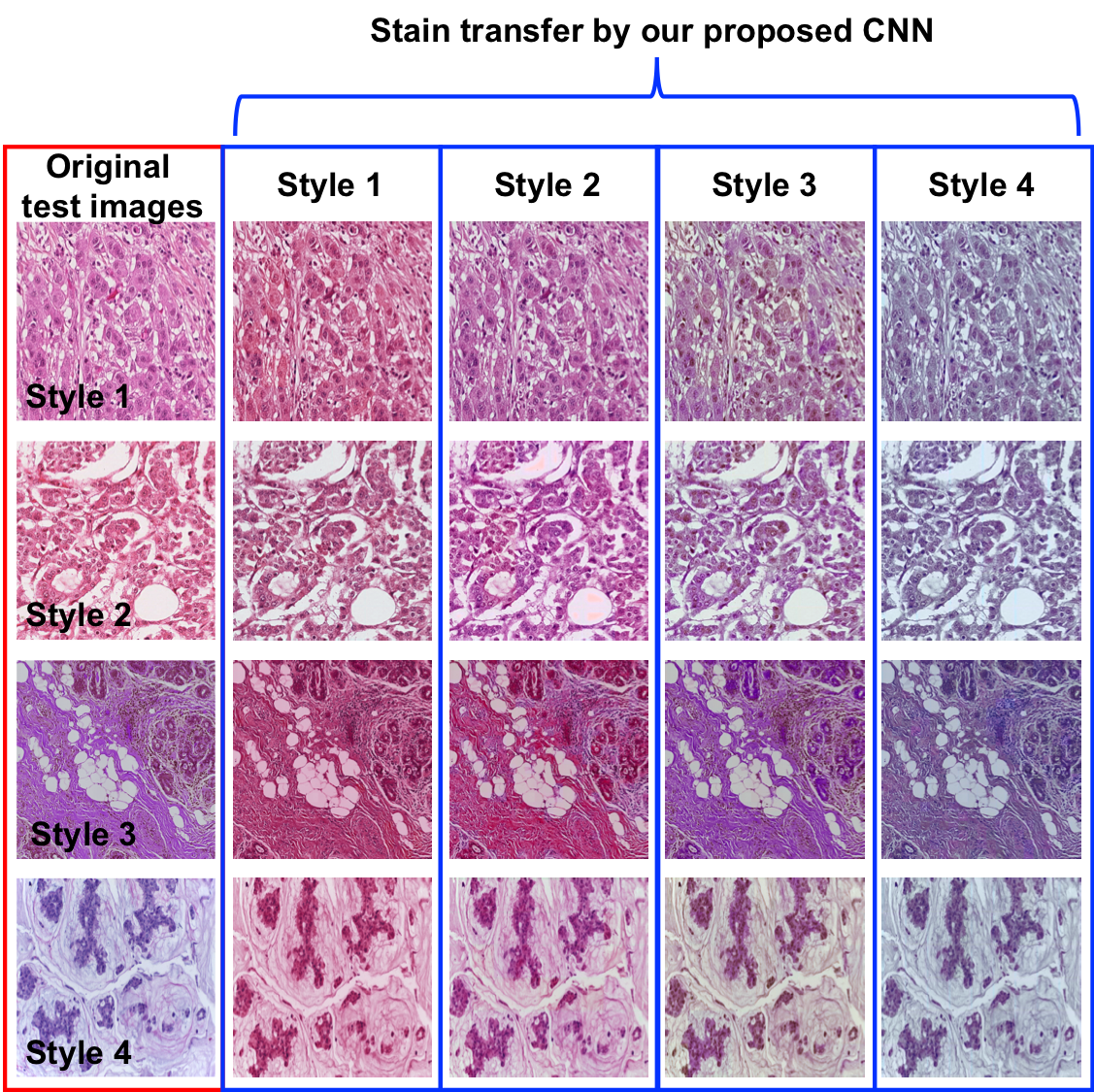}
\end{center}
   \caption{Examples of H\&E stain variantion in breast tissue in the left-most column.  The four columns to the right of this show examples stain transfer using our proposed CNN.}
\label{fig:variation}
\end{figure}

Given the scale and detail of the data, automated Computer Aided Diagnosis (CAD) technologies provide an intriguing alternative that is able both to utilize both, significantly larger quantities of data, and correlations between data that exist across different scales and different spatial locations.  We restrict the following discussion to the particulars of  breast cancer.  

From a diagnostic perspective, CAD technologies could provide reliable, quantitative metrics for grading the aggresiveness of a tumor, which is traditionally based in the Nottingham Grading System.  A weakness in the current grading system is that it is a qualitative score (low, moderate, strong) limited by 1) the medical practitioner's own biases 2) his inability to completely review massive amounts of histological data.  A CAD-based method would standardize the criteria used for identifying the aggresiveness of a tumor, while also providing improved medical recommendations using information across different spatial-scales that a human would not be able to interpret.

\subsection{Our Contributions}

We develop a three-step framework for quantitative and unsupervised learning on the breast cancer H\&E stained images. 

The first involves normalizing the stain appearance of H\&E images to a fixed reference stain appearance.  For this purpose, we train an optimized convolutional neural network on pools of images, each with a single stain style.  The network learns to correlate spatial featues in each image with their corresponding color characteristics.  Each network can then be used to recolorize grayscale images into its learned stain style.

After stain normalization, we train a U-net type neural network to segment out nuclei from the H\&E  images.  We train the segmentation network on a publicly available dataset and then show that stain normalization allows for improved numerical segmentation results.

In the final step of our framework, we segment cell nuclei in stain-normalized breast cancer histology images, and then perform classification of these nuclei using an unsupervised approach that has its roots in the Information Generative Adversarial Network (InfoGAN) architectures developed by ~\cite{cellinfogan}.  This architecture is similar to the basic GAN architecture with two additions: (1) A Lipschitz gradient penalty and the use of the approximate Wasserstein distance in the loss function (2) A mutual information loss term that encourages the generator to learn representations that the discriminator can then cluster into different classes.

\section{Problem Statement \& Related Work}

\subsection{Stain Normalization}

\begin{table}
\begin{center}
\begin{tabular}{|l|c|}
\hline
Dataset & Contents \\
\hline\hline
MITOS-ATYPIA-14 & Scanner A: 40,000 images \\
&  Scanner H: 40,000 images \\
 &  \small{40x 224x224 pix} \\
\hline
BACH ICIAR 2018 & 4 sets of 50,000 images total\\
& \small{40x, 224x224 pix} \\
\hline
van de Rijin \& West lab & 4 sets of 50,000 images total \\
& \small{40x, 224x224 pix} \\
\hline
Nuclei Segmentation& 2268 images \\
Benchmark & \small{40x, 144x144 pix} \\
\hline
\end{tabular}
\end{center}
\caption{Breast Histology Datasets.  The first three datasets were used for stain normalization.  The fourth for training nuclei segmentation.  The third for performing cell classification.}
\end{table}

Figure 1 shows examples of the variation in the H\&E staining outcome in the left-most column.   Our goal for the stain normalization is to recalibrate the color histogram of any histological image to that of a given reference image.  Prevous studies have shown that this stain normalization step improves the performance and reliability of downstream algorithms, in our case, nuclei segmentation and classification ~\cite{svdstain}.  

Variation in stain appearance can be due to a large variety of factors, such as variation in temperature, protocol, stain amount, and microscope parameters, that are very difficult to carefully control across different labs and even by the same scientist operating in different sessions. Examples of stain variability are shown in Figure ~\ref{fig:variation} .  Stain variability appears to have a largely detrimental effect on CAD algorithms that are not trained to deal with variation in stain color, with diminished performance in tasks such as segmentation and classification ~\cite{deconv}.  
We train a style transfer network (detailed below) on a global pool of images collected from various online competitions as well as from a lab at Stanford.  The totality of collected images is shown in Table 1.  We confine our training to 40x images that are cropped to regions of size 224x224, with horizontal and vertical shifts applied to augment data.

There is a rich and deep history of techniques that have been utilized for stain normalization.  A widely used method is based on Singular Value Decomposition (SVD) of the color space of a set of images~\cite{deconv}.  The two vectors with the largest singular values tend to correspond to the colors of the Hematoxylin and Eosin stains.  After obtaining these two vectors, we can use them as a color basis function for a second image, thus transforming them to the same color space.  This method, however, only takes into account the global color information of an image, and disregards the fact that color information corresponds locally to features in the image ~\cite{neuralstain}.  As such, it can be prone to errors as we illustrate in Figure 3.  More recently, there have a been a host of neural network based techniques for stain normalization, such as Variational AutoEncoders ~\cite{staingan}, deep Gaussian Mixture Models ~\cite{staingan}, and GANS~\cite{adversarialstain} that aim to address the aforementioned issues.  Our approach is similar to these latter techniques.  We propose to use our style transfer network as a conceptually simple but highly powerful technique for stain normalization, that learns to re-colorize images in a feature- based manner.
\begin{figure}[t]
\begin{center}
   \includegraphics[ scale=0.28]{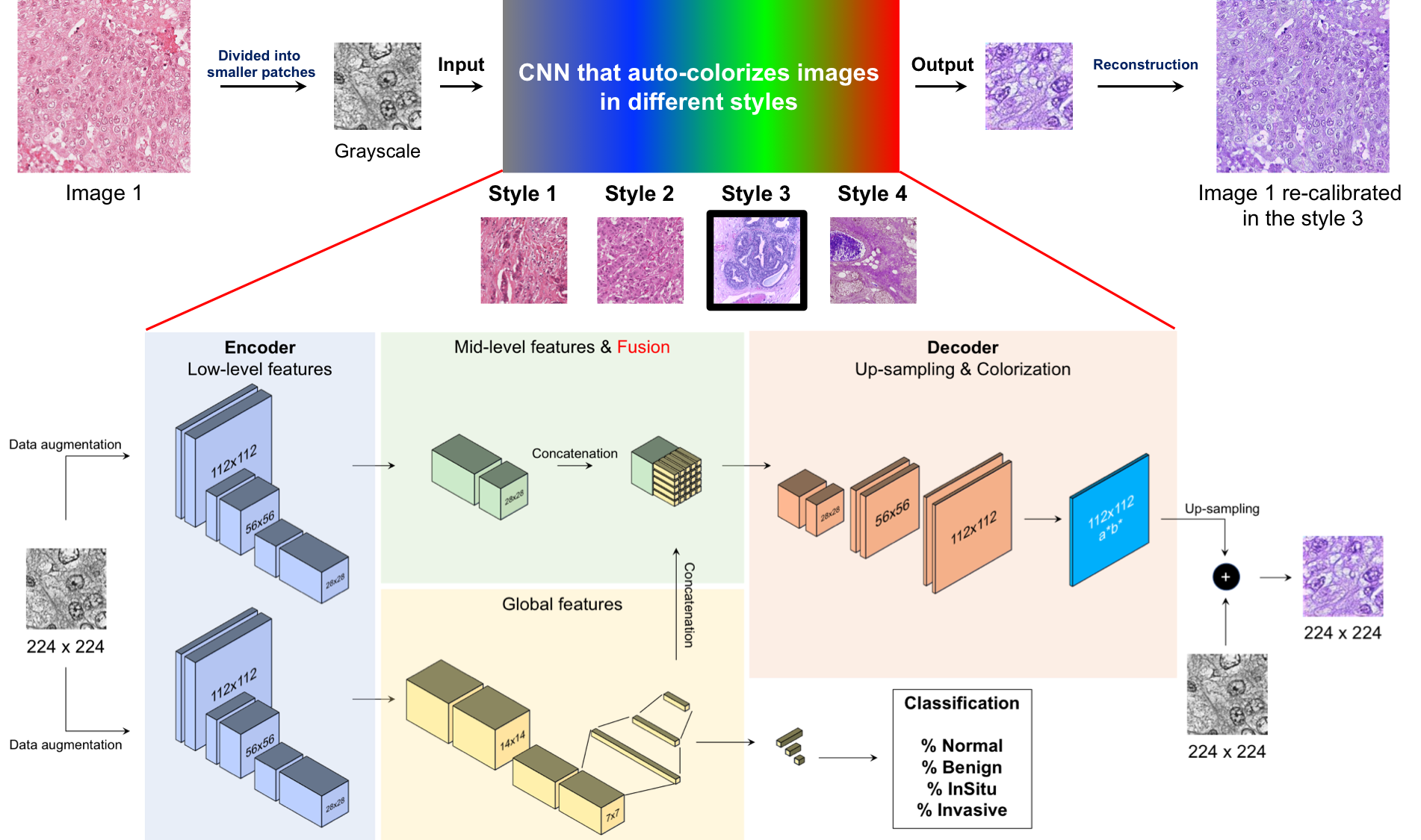}
\end{center}
   \caption{Stain transfer pipeline using CNN.  Our network recalibrates the stain style of image 1 to that of style 3.}
\label{fig:colorization}
\end{figure}

\subsection{Nuclei Segmentation}
Nuclei segmentation refers to the task of labeling every pixel in a H\&E image as either background, nuclei, or nuclear boundary.  Visually, the nuclear boundary is distinguishable by the fact that it heavily absorbs hematoxylin dye, and hence its typically ellipsoid shape and darker purple color are distinct from the background in H\&E images.  We train a neural net to perform this segmentation task using the Weebly nuclei dataset (Table 1), which consists of 2268 144x144 pixel H\&E crops with pathologist annotated nuclei.  We withhold 256 images for later testing purposes.  We perform rotation and inversion operations to augment the dataset.  Five percent of this dataset is also used for validation.  

Existing nuclei segmentation algorithms can be sorted into several classes: thresholding, region growing, level sets, k-means, and graph cuts ~\cite{cellseg}. All of these models have the disadvantage of having a large number of hyperparameters that need to be carefully tuned to match the nuclear shape, color, and characteristics of surrounding tissue, in order to achieve good performance.  More recent deep learning based models have combined deep feature learning with heuristic cellular features ~\cite{beyond} to yield highly optimized results, albeit ones that are not usually generalizable.  Our approach is quite similar to that of Cui et. al. ~\cite{contour}  We use data augmentation to train a U-net-like network end to end to perform pixel-wise classification of nuclei.

\subsection{Cell Classification}

Our aim for classification is to classify segmented nuclei (and hence cells) based on visual appearance in an unsupervised fashion by utilizing semantic features in the image, such as shape, nuclear density, and color.  There are numerous types of cellular nuclei in the image, including fibroblasts, lymphocytes, erythrocytes (lacking nuclei), epithelial, and cells in various stages of mitosis.  For our ground truth labels of cell types, we  sort the cells into three classes based on visual appearance: (1) epithelial and myo-epithelial cells which are large, and can have sparse, fragmented nuclei (2) lymphocytes, which have very spherical, dense darkened nuclei, andl (3) fibroblasts, which have long and thin nuclei, depending on the stage of the cell.  In total we have 644 type 1 nuclei, 308 type 2 nuclei, and 145 type 3 nuclei.  Examples of all three nuclei types can be seen in Figure, in the ground truth images.

We utilize `benign' breast histology images from the van de Rijn \& West lab at Stanford to train our cell classification architecture (Table 1).  We utilize 197, 350x350 pixel images to extract 1177 nuclei, which are segmented and resized to 40x40 images.  

Our work here borrows from the approach of Hu et. al. ~\cite{cellinfogan} to use an InfoGAN architure for unsupervised cell sorting.  There is a great deal of other related work in patch-level representation of histological tissue using deep learning techniques ~\cite{basal}, and fewer works on learning single-cell representations ~\cite{autoenc}.  One interesting related approach is to use dictionary learning to learn a cell image basis set of "cell words", from which one can construct any cell image in the dataset.  ~\cite{words}
\begin{figure}[t]
\begin{center}
   \includegraphics[ scale=0.3]{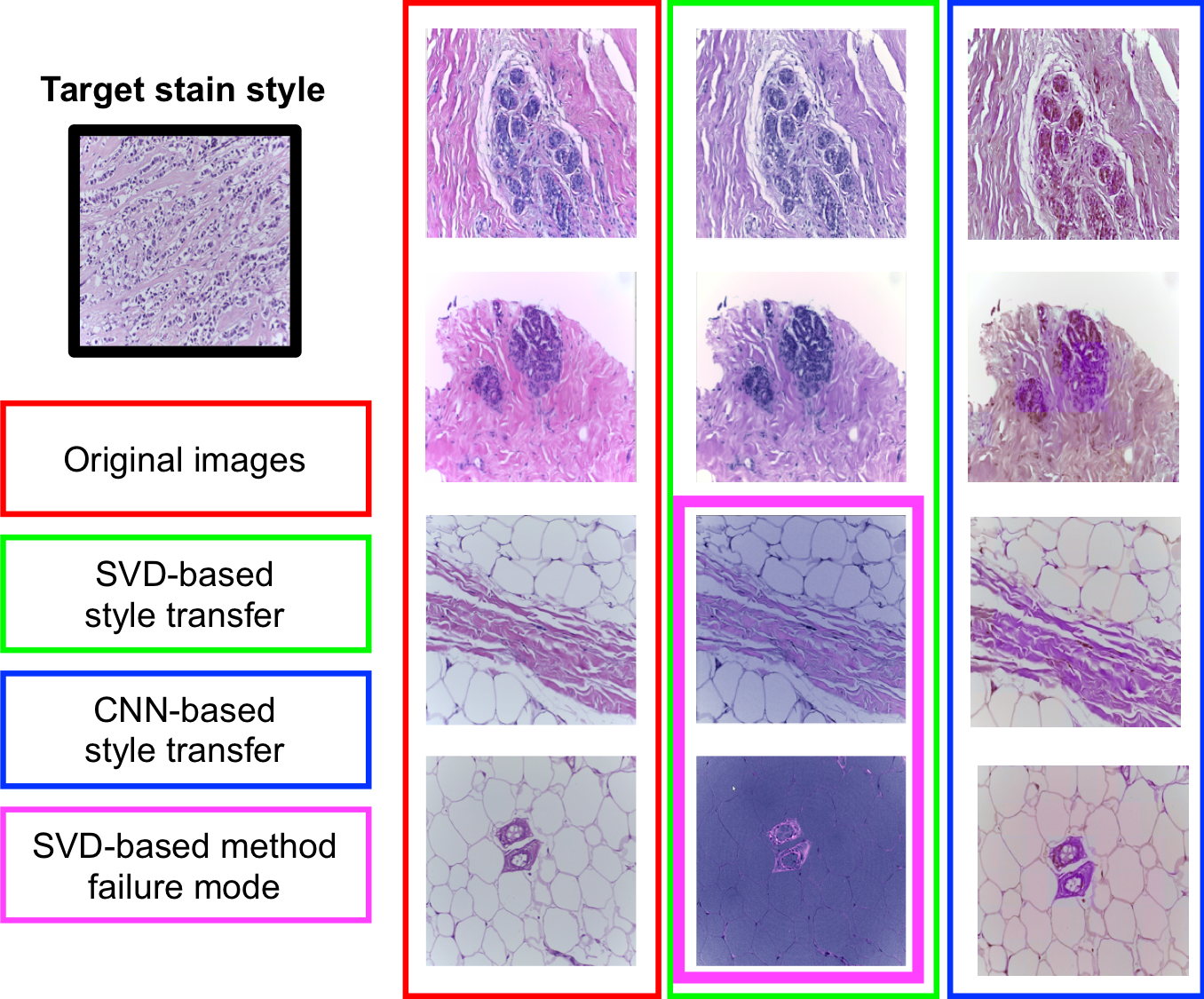}
\end{center}
   \caption{Comparison of the results of our CNN stain transfer with the baseline SVD method.  Images in the pink box show a typical failure mode for the SVD method, where the white background color is mistakenly colorized to the hematoxylin nuclei color.}
\label{fig:colorization}
\end{figure}

\begin{figure*}
\begin{center}
 \includegraphics[ scale=0.23]{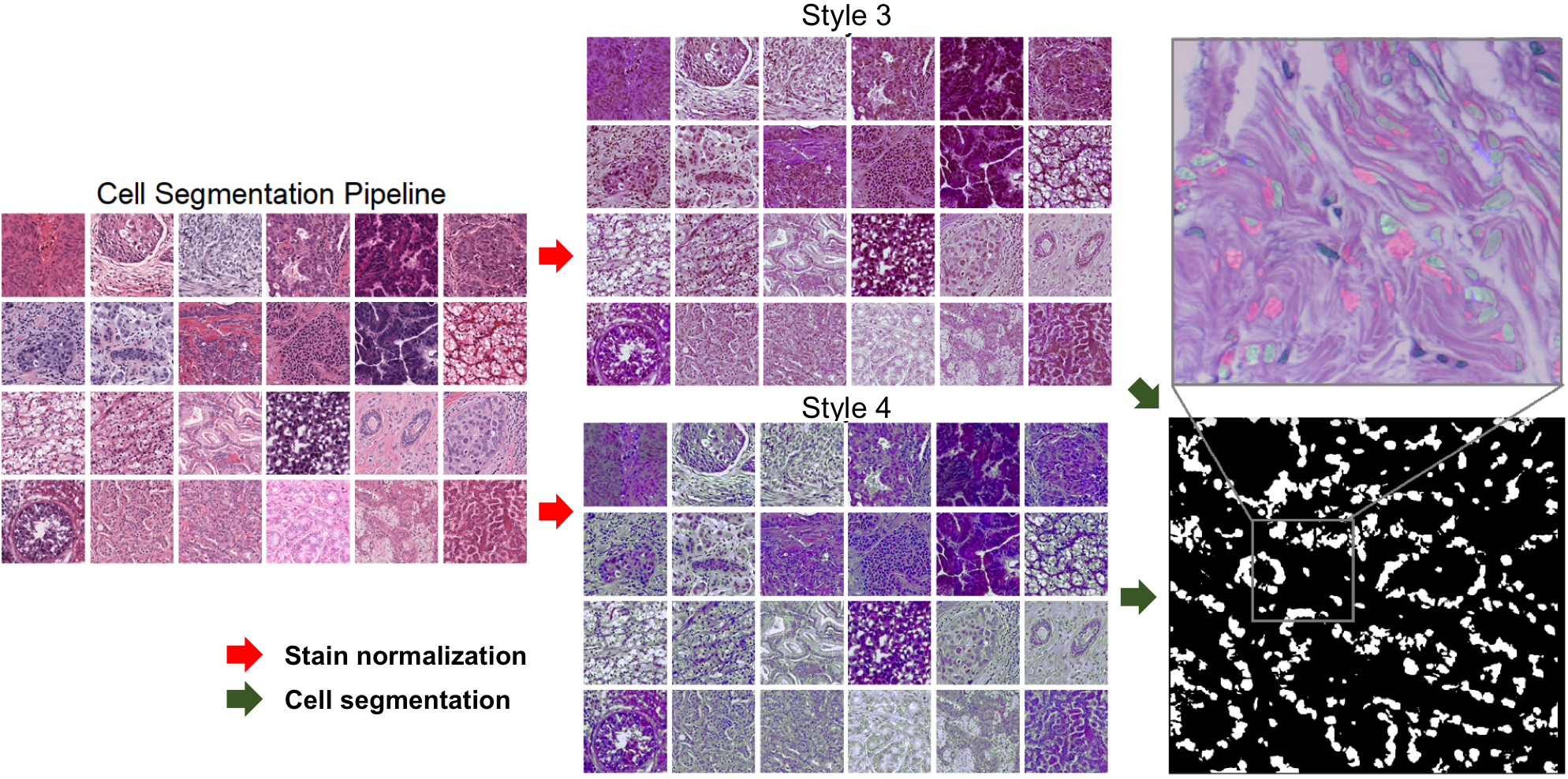}
\end{center}
   \caption{Illustration of the pipeline for our cell segmentation procedure, where a collection of images is first stain normalized.  A binary segmentation map is generated with pixels either within nuclei, or outside of it (lower right image).  The upper right image shows the difference in the segmentation between the two stain normalized datasets shown here.  Red and green colors indicate nuclei segmented in only one of the datasets, while blue shows nuclei segmentation common to both (upper right).}
\label{fig:short}
\end{figure*}

\section{Technical Approach \& Method}
\subsection{Stain Normalization}
We approach the stain normalization problem as a color re-calibration problem. We built a CNN that takes grayscale images as an input and outputs color components of images. In 2016, Iizuka et al ~\cite{style} proposed a novel approach where a network was end-to-end trained using pixelwise mean squared error (MSE) loss along with classification loss to capture semantic information in colorizing images. Our network is primarily inspired by that architecture.

 The model consists of three parts: the encoding, classification, and colorization units. The architecture is depicted in Figure 2. 

\begin{enumerate}
\item The encoder serves as the low-level feature extractor and processes 224x224 grayscale histology images and outputs a 28x28x256 feature representation as the ‘mid-level’ representation. It uses 8 convolutional layers with 3x3 kernels. Padding is used to conserve the size of input. Furthermore, instead of using max-pooling, convolutions of stride of 2 are used to reduce the number of computations required while preserving spatial information.
\item The classification is done by a global feature extractor which has 4 convolution layers with final fully connected layers to classify the histology images
into their respective categories (Normal, Benign, Ductal Carcinoma In-Situ and Invasive). The output of the global feature extractor is a 256-dimensional vector representation, which is concatenated with the output of the encoder for colorization. 
\item The colorization is done by a decoding unit that takes the output of the concatenation layer and applies a series of convolutional and up-sampling layers.
\end{enumerate}

ReLU was used as a transfer function and batch normalization are performed at each layer to allow for more stable training.  The network was trained with the Adadelta algorithm ~\cite{adadelta} with batch size =64, learning rate = 1.0, rho = 0.9, eps =1e-6, and no L2 penalty.  We used varied ratios of classifcation loss to total loss from 1/30 to 1/300. 

The network has two sources of loss.  One from pixelwise mean squared error (MSE) loss in reconstructing the colorized image.  The second loss arises from classification loss in sorting images between different tissue types as our dataset consists of labeled images from four different breast cancer development stages - normal, benign, ductal carcinoma in-situ, and invasive.  We find the classification loss especially to be helpful in training the low-level feature network, which is farther upstream from the MSE loss, and otherwise does not receive enough gradient to be properly trained.  In fact, we observe artifacts in some colorized images without the classification loss (refer to Appendix).

\textbf{Evaluation:}  We first qualitatively compare the results of our neural stain transfer technique with a highly ubiquitous technique based on singular value decomposition~\cite{deconv}.  We've discovered this model has several failure modes that are intrinsic to the way it functions (described in Technical Approach, see Figure 3), and we show that our neural stain transfer architecture is able to side-step these pitfalls.

Secondly, we evaluate the effectiveness of our stain normalization by comparing its performance on our downstream task, nuclei segmentation, which is dicussed in the following section.

\subsection{Nuclei Segmentation}

Our nuclei segmentation network is a U-net like structure consisting of layes of 4  layers of downsampling and 4 symmetrical layers of upsampling, bringing images of original pixel size 144x144 down to 512 channels of 9x9 images and then back up to the original image size.  Each layer consists of 2 convolution operations and a maxpool operation followed by ReLu.  Skip connections concatenate the equivalently sized downsampled images with each upsampling layer.  During training we use an Adam Optimizer with learning rate of 1e-4.

\textbf{Evaluation:}  
We evaluate nuclear segmentation based on several numerical metrics.  We compute pixel-wise sensitivity, specificity, Dice score and Jaccard similarity scores.  The definitions are stated in the Appendix.

\subsection{Cell Classification}

\begin{figure}[t]
\begin{center}
   \includegraphics[ scale=0.3]{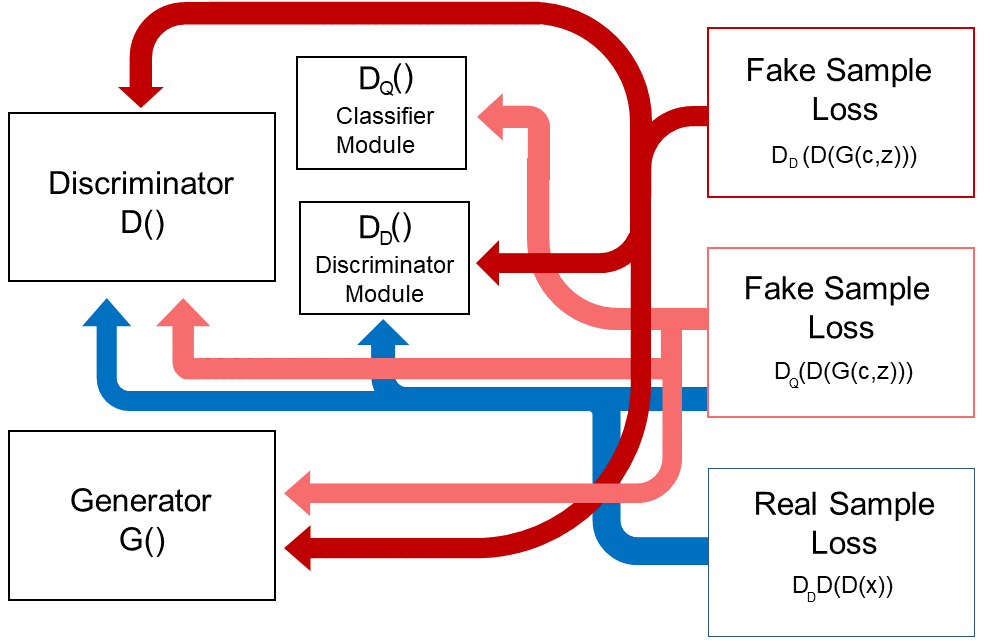}
\end{center}
   \caption{Gradient flow in our GAN-based unsupervised classification network.}
\label{fig:colorization}
\end{figure}

After normalizing stain color, we use the HistomicTK Python API to perform cell segmentation of 1177 nuclei via the Laplacian of Gaussian blob identification algorithm.  The results are not optimal in 100\% of cases, so corrections to the algorithm's segmentation were made by hand as needed.  Although the neural nuclei segmentation provides superior results, it was not developed in time in the early stages of the project for use.  Labeling of the cells into one of three classes was done by the authors in consultation with a biology postdoctorate scholar at Stanford.  

We use an InfoGAN type architecture ~\cite{cellinfogan} that learns to classify image classes and stores semantic information in an additional latent variable termed $c$.  In this model, the disciminator $D$ is connected to two modules: a standard adversarial discriminator $D_D$ and an auxillary network $D_Q$ that takes as input generated images $G(z,c)$, and outputs the inferred class $c$ of the image (Figure 5).  It is explicitly optimized to maximize its mutual information $I$ with the latent variable $c$. The random vector $z$ is the classic noise input of GAN-type networks.  The mutual information loss is given by:
\begin{equation}
I(c,G(z,c)) \geq \mathbb{E}_{z\sim p(z),c\sim p(c)}D_Q(D(G(z,c)))
\end{equation}
where the term on the right is a lower bound for the actual loss and is actually utilized in gradient descent.  In our current run settings, $z$ is a 16 dimensional random variable, $c$ is a 3 class categorical variable.  

Finally our GAN architecture makes use of the Lipschitz penalty ~\cite{wasgan} and Wasserstein distance.  In the original Wasserstein GAN paper ~\cite{wasgan1}, it is shown that the Wasserstein distance has certain benefits that make it a more suitable metric when training a model to reproduce a data distribution, compared to the Jensen-Shannon divergence of standard GAN models.
\begin{equation}
\text{Lipschitz penalty} = \mathbb{E}_{x\sim P_x}(\lVert\nabla_{x} D_D(D(x))\rVert-1)
\end{equation}
\begin{equation}
\text{Wasserstein Dist.} \approx \mathbb{E}_{x\sim P_r}D_Q(D(x)) - \mathbb{E}_{z\sim P_z} D_D(D(G(c,z)) 
\end{equation}
These revisions to the loss lead to improved smoothness of the loss function and interpretability of the discriminator loss (which is shown to correlate with generated image quality).  We train with the following parameters: learning rate of $D_Q$ network is 2e-4, learning rate of $G$ network is 1e-4, and learning rate of $D_D$ network is 1e-4.

\textbf{Evaluation:}  We propose several different evaluation metrics. Our first evaluation metric is thus to verify that the model is able to sort a cell population into the three basic classes mentioned above.  We evaluate both the sensitivity and precision of each class labeling and we also utilize the F1 score, or the harmonic mean of precision and sensitivity.  Unlike the standard mean, this quantity is diminished significantly if either the precision or sensitivity is low, and thus serves as a single indicator of the test's performance.  

As comparisons, we evaluate a model based on L2 distance with k-means clustering of 3 clusters, representing each image as a squeezed 3072-length vector.  We also compare L2 distance based k-means clustering using features of length 2048 from the ``average pooling" layer of a pre-trained Res-Net 152 network.

A second evaluation metric is the confusion matrix of classification, which helps us understand which classes the model is confounding together.

\section{Results \& Dicussion}
\subsection{Stain Normalization}

We qualitatively compare our CNN stain transfer results with results from the common-practice SVD technique. Figure 1 shows numerous stain converted images by our network.  Figure 3 shows that our CNN technique is able to overcome difficulty with normalizing white spaces in H\&E images (third and fourth rows of Fig. 3) that the SVD technique encounters. 


We also performed quantitative performance comparisons where we used 100 test images from the BACH dataset (Table 1).  In the table below we first compare the L2 Euclidean distance between images that are stain normalized to a target style using the SVD method, the CNN-method with no classification loss, and the CNN-method with classification loss.  We see that pixel-wise, our method (last row in the table) produces an output that is visually closer to the targeted image.  
\begin{table}[htbp]
\centering
\begin{threeparttable}[hb]
\begin{tabular}{|c|c|c|c|}
\hline
\multirow{3}{*}{\footnotesize{Model}} &  \multicolumn{1}{|c|}{\footnotesize{Distance Metric}} &\multicolumn{2}{|c|}{\footnotesize{Color Histogram}}   \\ \cline{2-4}
& \footnotesize{Euclidean} & \footnotesize{Correlation} & \footnotesize{Hellinger}  \\ 
& \footnotesize{(0 is ideal)} & \footnotesize{(1 is ideal)} & \footnotesize{(0 is ideal)}  \\ \hline
\footnotesize{SVD-based}& \multirow{ 2}{*}{1.25} & \multirow{ 2}{*}{0.18} & \multirow{ 2}{*}{0.83}  \\
\footnotesize{stain transfer}  &  &  &  \\ \hline
\footnotesize{CNN-based}  & \multirow{ 3}{*}{1.01} & \multirow{ 3}{*}{0.43} & \multirow{ 3}{*}{0.68}  \\
\footnotesize{stain transfer} &  &  &  \\
\footnotesize{(MSE Loss)} &  &  &  \\ \hline
\footnotesize{\textbf{CNN-based}} & \multirow{ 3}{*}{0.98} & \multirow{ 3}{*}{0.48} & \multirow{ 3}{*}{0.58}  \\
\footnotesize{\textbf{stain transfer}}  &  &  &  \\
\footnotesize{\textbf{(MSE+class. loss)}} & &  &  \\ \hline
\end{tabular}
\begin{tablenotes}[para,flushleft]
\end{tablenotes}
\end{threeparttable}
\end{table}

We also compare the color histograms of the pre- and post- normalized images.  We find that our best performing model on average has a significantly higher color histogram correlation (averaged between RGB channels) with the color histogram of the targeted stain.  We believe that this highly positive result is due in part to the fact that the SVD method made significant errors in normalizing background white colors throughout our test dataset.  The increase in the color histogram correlation between our models with and without classification loss is also likely due to the artifacts present in the latter model (see Technical Approach for dicussion). Finally, the Hellinger distance between the color histograms also reveals a similar trend, where our CNN network with classification loss achieves the best performance.

\subsubsection{Impact of Stain Normalization on Cell Segmentation}
Next we characterize the performance of the stain normalization on cell segmentation which is critical for the ultimate step of cell classification. We expect models where both training and testing images are stain normalized should perform better than both models with (1) no normalization of training or tests and (2) training is normalized but testing is not normalized.

We observe that indeed our CNN stained model, which is both trained and tested on stain normalized images, achieves improved sensitivity of 9\% over the the baseline model, which was tested and trained on imags from a wide variety of styles.  This is shown in the table below.  We also note a sensitivity boost compared to the SVD model and the SVD model also attains a ~3\% boost in performance compared to the baseline.  However, we note that we experience a substantial drop in specificity, which was also visually obvious from the fact that the CNN model was overly-aggresive in its nuclei segmentation.  

\begin{table}[htbp]
\centering
\begin{threeparttable}[hb]
\begin{tabular}{|c|c|c|c|c|}
\hline
\multirow{2}{*}{\footnotesize{Model}}& \multirow{2}{*}{\footnotesize{Sensitivity}} & \multirow{2}{*}{\footnotesize{Specificity}} & \footnotesize{Dice}  & \footnotesize{Jaccard} \\
& & & \footnotesize{Score}  & \footnotesize{Similarity} \\ \hline
\footnotesize{Baseline}& \multirow{ 2}{*}{50.0} & \multirow{ 2}{*}{88.2} & \multirow{ 2}{*}{29.3} & \multirow{ 2}{*}{17.2}\\
\footnotesize{Un-norm.} &  &  &  &\\ \hline
\footnotesize{SVD}  & \multirow{ 2}{*}{53.3} & \multirow{ 2}{*}{82.2} & \multirow{ 2}{*}{30.6}  & \multirow{ 2}{*}{18.1}\\
\footnotesize{Norm.} & &   &  & \\ \hline
\footnotesize{\textbf{CNN}} & \multirow{ 2}{*}{59.3} & \multirow{ 2}{*}{63.3} & \multirow{ 2}{*}{33.0}  & \multirow{ 2}{*}{19.7}\\
\footnotesize{\textbf{Norm.}}& &  &  & \\ \hline
\end{tabular}
\begin{tablenotes}[para,flushleft]
 \footnotesize{all values give in (\%)}
\end{tablenotes}
\end{threeparttable}
\end{table}

Overall however, taking into account the trend in the Dice score and Jaccard similarity, we conclude that stain normalization has a positive effect on the nuclei segmentation results.  We also anticipate that the SVD stain-normalization model would experience a larger performance improvement were it not for the background artifacts present in the images.  


\subsection{Cell Classification}
\begin{figure}[b]
\begin{center}
 
  \includegraphics[ scale=0.23]{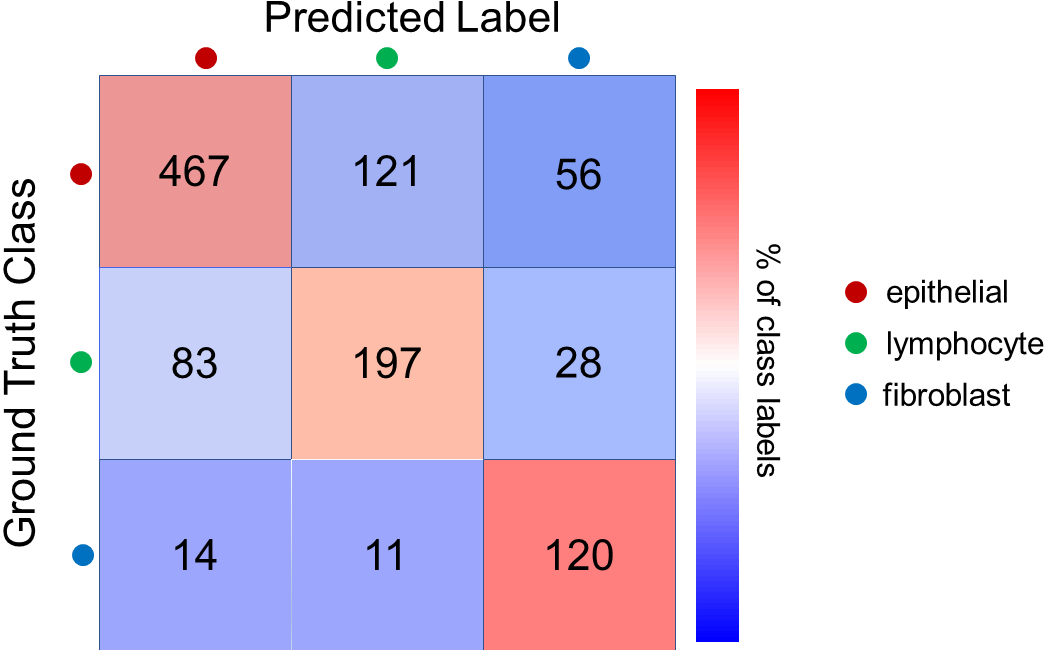}
\end{center}
   \caption{Style transfer pipeline using CNN's.  Global features of image 2 get fused with low-level features of image 1 to colorize image 1 in the style of image 2.}
\label{fig:colorization}
\end{figure}

\begin{figure*}
\begin{center}
 \includegraphics[ scale=0.5]{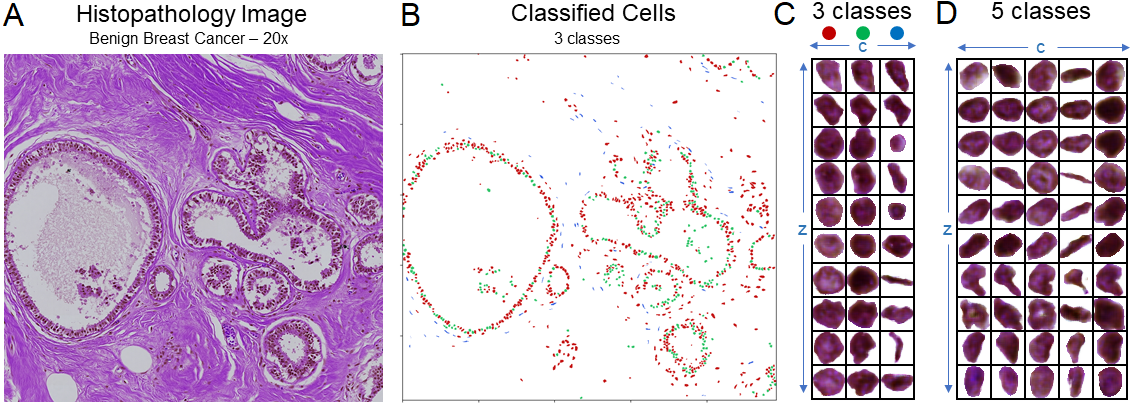}
\end{center}
   \caption{Figure A shows an H\&E image of benign breast cancer tissue.  B shows the nuclei labels predicted by our unsupervised classification algorithm in the 3 class case.  C and D show generated images from our GAN-based unsupervised classification algorithm.  Each column shows a set of nuclei with different class labels $c$. Each of the classes is also labeled with a color.   Each row shows a different initialization of latent variable $z$, where the latent variable $z$ is held constant across all columns. }
\label{fig:short}
\end{figure*}

The results from the segmented and then classified nuclei are shown in Figure 7. Figure 7A shows the original H\&E image of a benign breast cancer sample.  Figure 7B shows the segmented nuclei along with the color of the model's predicted class of the nuclei, for 3 types of cells- epithelial, lymphocytes, and fibroblasts.  As expected, epithelial cells (red) are mostly labeled as occuring in circular lobule and duct like structures.  Biologically, these structures are composed of layers such cells.   Long, thin fibroblasts (green) are scattered throughout the surrounding stromal tissue, and lymphocytes (blue) are interspersed between the epithelial layers and in cluster-like surrounding structures.  

Figures 7C and 7D show generated images for 3 classes and 5 classes that were learned by our cell classification network after 70,000 iterations of training.  In the case of 3 classes, we see a distinct separation of learned representations into 3 morphologically distinct cell types.  The first type of cell nuclei (corresponding to epithelial), tends to be larger and have a light purple colored nucleus with more darkly colored nuclear fragments inside.  The second class of cell (corresponding to lymphocytes) has a more uniform, darker nuclei compared to the first class.  The third class of cell (corresponding to fibroblasts), has a long and thin nucleus.  

Comparison of the fake, generated images with the real, ground truth images is shown in Figure 8.  It is apparent that our model learns quite a plausible representation of each class of nuclei.  There are two primary differences that are observable: 1) The difference between the epithelial and lymphocyte class is less apparent in our model.  There are images in either class that could plausibly also belong to the second class.  2) The fibroblast class in our model contains nuclei with a much greater variety of shapes than in the ground truth class, where almost all fibroblasts have a similar lengthy, spindle shape.

In the case of 5 classes (Fig. 7D), we can again see the emergence of the 3 classes we discussed above in columns 3,4, and 5.  Columns 1 and 2 appear to be variants of the epithelial and fibroblast classes.  It is interesting to note that although, visually, the class differences between the first two columns are not apparent to a human observer, the model is extremely confident about its predictions of most class labels (see Figure 10 in the appendix).  Thus there is some human-imperceptible aspect of the generated images that the discriminator model is able to utilize to confidently make a prediction about the class label of the image.

Finally, we note that the model learns a representation of the rotational degree of freedom encoded in the latent $z$ variable that is apparent in both the 3 class and 5 class models.  It can be observed that images with a shared $z$ variable almost always have the same rotational orientation.  We believe this phenomenon is perhaps enhanced by the fact that while training, the dataset is augmented with rotational variations of each image.  Certain aspects of shape, on the other hand, are encoded by both the $c$ and $z$ variables.

\begin{figure*}
\begin{center}
 \includegraphics[ scale=0.5]{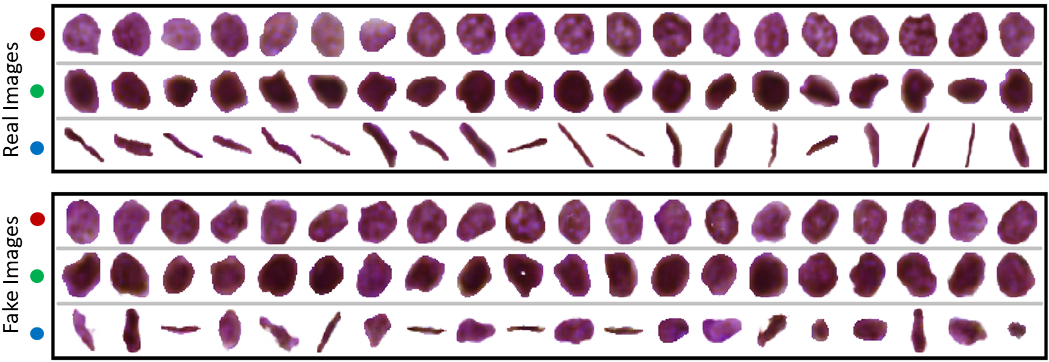}
\end{center}
   \caption{The top figure shows real images of each class from the ground truth dataset.  The bottom figure shows images learned by the generator in our unsupervised classification model after training for 70,000 iterations. }
\label{fig:short}
\end{figure*}

\subsubsection{Numerical Evalulation}
\begin{center}
\begin{tabular}{|c|c|c|}
\hline
 Model & Precision (\%) & Sensitivity (\%)\\
\hline
\footnotesize{\textbf{InfoGan (Our Model)}} & \footnotesize{82.8/59.5/58.8}  & \footnotesize{72.5/64.0/82.8} \\
\hline
\footnotesize{ K-means} &  \footnotesize{66.9/35.6/45.3} & \footnotesize{35.2/46.8/63.4} \\
\hline
\footnotesize{ResNet-152 features K-means}  & \footnotesize{72.1/38.4/39.5} & \footnotesize{35.6/52.9/77.9}  \\
\hline
\end{tabular}
\end{center}

The table above gives statistics regarding performance for each the classes in the format: \textit{class1 / class2 / class3 (epithelial / lymphocyte / fibroblast)}. 

We observe that the model shows strong sensitivity for each of the three classes, with the highest sensitivity for the fibroblast class at 82.8\% and the lowest for the lymphocyte class at only 64\%.  On the other hand, the fibroblast class also shows rather low precision at only 58.8\%, indicating that some of our model's success with that class is due to over- labeling.  The same can be said for the lymphocyte class, with a precision of only 59.5\%.  The lymphocyte class is inherently challenging because it strongly resembles and can bear some overlap with the largest class - epithelial cells.

The confusion matrix for our model (Figure 6) demonstrates the same point, that the largest class overlap from mislabeling is due to epithelial-lymphocyte confusion.  

We compare our model against two baselines models: k-means on the pool of image vectors and also k-means on ResNet-152 features, each with 3 clusters.   Notably, both versions of the the k-means algorithm suffer from dramatically reduced performance on epithelial cells compared to our model, with sensitivity of only around 35\% compared to the 72.5\% we observe.  

Visualization of the 2 largest principle components (Appendix- Figure 9) for both the entire image vector and for the features from ResNet also confirm that it is highly difficult to disentangle the epithelial and lymphocyte nuclei types.  It is surprising that the pretrained ResNet-152 did not yield an improvement in the separation of those two class types.  One hypothesis is that the two largest components of the principle component analysis tend to correspond to high-level morphological changes in the nuclei shape, but are not representative of finer grained image features, such as nuclear density, which can help differentiate between epithelial and lymphocyte cells.  

Finally, we summarize the sensitivity and specificity results using the F1 score and Accuracy in the table below:
\begin{center}
\begin{tabular}{|c|c|c|}
\hline
 Model & F1 & Accuracy\\
\hline
\footnotesize{\textbf{InfoGan (Our Model)}}& \footnotesize{77.3/61.6/68.76} & 66.6 \%\\
\hline
\footnotesize{ K-means} & \footnotesize{46.1/40.4/52.8}  & 44.3 \% \\
\hline
\footnotesize{ResNet-152 features K-means}  & \footnotesize{47.7/44.5/52.4} & 47.2 \% \\
\hline
\end{tabular}
\end{center}

We conclude that our model outperforms both types of k-means models by a significant margin on accuracy alone, while also enabling significantly improved sensitvity and specificty (reflected in the F1 score) for epithelial and lymphocyte type cells, probably by identifying low level image features such as nuclear density and fine-scale color changes that are characteristic of those nuclei types.

\section{Conclusions}
In this work, we developed an entire pipeline for the stain normalization of H\&E stained histopathology images, segmentation of nuclei, and unsupervised classification of nuclei, based all on deep neural network techniques.  We qualitatively demonstrate the effectiveness of our stain normalization algorithm, and then quantitatively show that it gives improved nuclei segmentation performance on an evaluation dataset.  We also demonstrate that our unsupervised cell classification model significantly outperforms various k-means based techniques on unsupervised classification.  

We would like to point out numerous opportunities for improvement in our present pipeline.  It is quite apparent that our unsupervised classification model does not learn entirely ``disentangled" semantic representations of image features between its various hidden variables.  Furthermore, the learned latent representation is one that is not always human interperatable, which poses a deep and ongoing topic of research ~\cite{ali}.

We identify two points here that may be promising future avenues of research: 

1) The model is highly confident in its classification predictions, even when incorrect (Figure 10 in Appendix).   The generated images are also highly realistic, indicating that the generator's adversarial game has converged to its desired result.  It thus appears that the discriminator's classification system has simply converged to a non-ideal result (from a human perspective).  One interesting possibly would be to ask the discriminator not only to return the classification label $c$, but also to estimate the real-valued latent variable $z$.  Doing so requires that it can reconstruct $z$ only from image information, thus requiring that $z$ be semantically meaningful.

 2) Even when trained to convergence, the model predicts numerous classes that are highly perceptually similar to a human observer.  This indicates that the discriminator is using some set of image feature different from what a human uses to classify images, and may give some insight into the differences between human and machine perception.  Another interesting possiblity would be to use a ``weaker'' or more high-level discriminator classifier network, thus requiring that the learned class $c$ correspond to image features that are easily recongnizable, and hence highly interpretable.

\clearpage

\textbf{Acknowledgements}
The authors greatly appreciate inspiration and support from Lukasz Kidzinksi, Magdalena Matusiak, and Korsuk Sirinukunwattana.  We also thank Magda for assisting our understanding of histological sections, and breast histology cell-types, and Korsuk for providing the segmentation codebase and technical support regarding it.

We would also like to thank CS231N staff for a great quarter, and particularly Carolyn Kim, Vincent Chen, Winnie Lin, Ajay Mandlekar, and Bingbing Liu for their assistance throughout the quarter.
{\small
\bibliographystyle{ieee}
\bibliography{egbib}
}


\end{document}